\pdfoutput=1
\documentclass[conference]{IEEEtran}
\IEEEoverridecommandlockouts
\usepackage{cite}
\usepackage{amsmath,amssymb,amsfonts}
\usepackage{algorithmic}
\usepackage{graphicx}
\usepackage{float}
\usepackage{orcidlink}
\usepackage{siunitx}
\DeclareSIUnit\knot{kn}
\usepackage{textcomp}
\usepackage{csquotes}
\usepackage{xcolor}
\begin{document}

\title{Design and Experimental Validation of an Autonomous USV for Sensor Fusion-Based Navigation in GNSS-Denied Environments}

\author{\IEEEauthorblockN{Samuel Cohen-Salmon  \orcidlink{0009-0006-4368-9516} \& Itzik Klein  \orcidlink{0000-0001-7846-0654}}
\IEEEauthorblockA{\textit{The Hatter Department of Marine Technologies,} \\
\textit{Charney School of Marine Sciences, University of Haifa, Israel}}}

\maketitle

\begin{abstract}
This paper presents the design, development, and experimental validation of MARVEL, an autonomous unmanned surface vehicle built for real-world testing of sensor fusion-based navigation algorithms in GNSS-denied environments. MARVEL was developed under strict constraints of cost-efficiency, portability, and seaworthiness, with the goal of creating a modular, accessible platform for high-frequency data acquisition and experimental learning. It integrates electromagnetic logs, Doppler velocity logs, inertial sensors, and real-time kinematic GNSS positioning. MARVEL enables real-time, in-situ validation of advanced navigation and AI-driven algorithms using redundant, synchronized sensors. Field experiments demonstrate the system’s stability, maneuverability, and adaptability in challenging sea conditions. The platform offers a novel, scalable approach for researchers seeking affordable, open-ended tools to evaluate sensor fusion techniques under real-world maritime constraints.
\end{abstract}


\section{Introduction}

Unmanned surface vehicles (USVs) are becoming essential tools across a range of maritime applications. These platforms have been deployed in domains including environmental monitoring, oceanographic data collection, port security,search and rescue, and offshore inspection \cite{9977269,10682271,10682411}. Depending on their mission profiles, USVs vary widely in size, architecture, propulsion, and endurance. From compact one-person deployable units to large, long-endurance ocean-going vessels, the diversity of USV designs continues to grow \cite{oceanalpha2023choosingUSV,navy2024largeUSVs}. Hull types such as monohulls, catamarans, and trimarans each offer distinct trade-offs in stability, payload capacity, and hydrodynamic efficiency \cite{9389152,9977269,9775277}.

Similarly, the choice of propulsion systems depends on the mission. Jet drives offer high maneuverability for patrol applications, while propeller-based configurations provide consistent performance for survey and research tasks \cite{9389085,10682271}. Increasingly, hybrid and renewable-powered USVs—using solar panels, wave energy, or sails—extend operational endurance with minimal supervision. These developments reflect a broader shift toward autonomy and sustainability in maritime systems \cite{10682411}.

A critical enabler of USV autonomy is their navigation system. Accurate and reliable navigation is essential not only for mission execution, but also for safety. Navigation systems typically combine multiple sensors, such as global navigation satellite systems (GNSS), inertial measurement units (IMU), Doppler velocity logs (DVL), and electromagnetic (EM) speed logs \cite{10244581,9775277}. These sensors work together through sensor fusion algorithms to estimate the vehicle’s state in real time, particularly when GNSS signals are unavailable.

In recent years, deep-learning algorithms are incorporated within the navigation algorithms \cite{cohen2024inertial,article}. One of the first applications of deep learning to underwater navigation was construction of missing DVL beams, which in turn enabled the estimation of the vehicle velocity vector \cite{yona2021compensating,cohen2023set,yona2024missbeamnet}. 

In USVs, deep learning has been applied to enhance inertial navigation systems and dead-reckoning accuracy, as well as to interpret sonar data for improved perception and environmental mapping \cite{turn0search9,turn0search15}. On the surface, USVs benefit from deep learning integration in navigation, guidance, and control pipelines, supporting operations in complex maritime environments. Reinforcement learning, in particular, has enabled USVs to learn adaptive navigation strategies through interaction with dynamic surroundings, offering a promising alternative to traditional model-based controllers \cite{turn0search3,turn0academia23}.

Other approaches, employ deep learning methods to adaptively estimate the process noise covariance in the inertial DVL fusion process \cite{or2023pronet,cohen2025adaptive,levy2025adaptive}.

Such algorithms rely on high-quality and diverse datasets for training and testing. In addition, most literature works employ simulation or semi-experimental approaches. In the latter, a single sensor is used to record actual data serving as the ground truth. The unit under test is a generated sensor created by adding a noise model to the ground truth (GT). These two gaps pose a challenge for researchers aiming to evaluate navigation algorithms under real-world conditions.

To address these limitations, we present the design and development of MARVEL — a modular AI research vessel for experimental learning. MARVEL is a cost-effective, fully autonomous USV built to support real-time sensor fusion experimentation and autonomous navigation research. Its modular architecture supports easy integration of multiple redundant sensors, including DVLs, EM logs, IMUs, and GNSS. The system was designed to be portable, seaworthy, and reconfigurable, enabling consistent performance across a range of sea conditions.\\

Unlike many commercially available platforms, MARVEL is tailored for high-frequency data acquisition, extended deployments, and affordable customization. It was developed using off-the-shelf components, open-source software, and a low-cost chassis modified for stability and robustness. Through iterative field testing, the platform demonstrated high reliability, maneuverability, and adaptability in both calm and challenging marine environments.

Once the system architecture was finalized, software development focused on implementing advanced navigation, guidance, and control algorithms tailored for real-world maritime conditions. The testing phase consisted of laboratory-based validation of electronic, control, and power subsystems, followed by real-world trials under progressively challenging sea states. These field tests provided critical insights into MARVEL’s performance, allowing for iterative refinements and system optimizations. 

Through a combination of modular design, open-source software, and cost-efficient hardware selection, MARVEL was developed to serve as a versatile and scalable research platform, supporting a wide range of sensor fusion experiments in autonomous marine navigation.

The rest of the paper is organized as follows: Section~\ref{sec:sow} presents the scope and design objectives of the platform. Section~\ref{sec:hardware} describes the system design, including hardware architecture and sensor integration. Section~\ref{sec:assembly_testing} details the assembly process, testing methodology, and experimental validation and Section~\ref{sec:conclusion} concludes this work.

\section{Core Design Principles} \label{sec:sow}

\begin{figure}[b]
\centering
\includegraphics[width=0.5\textwidth]{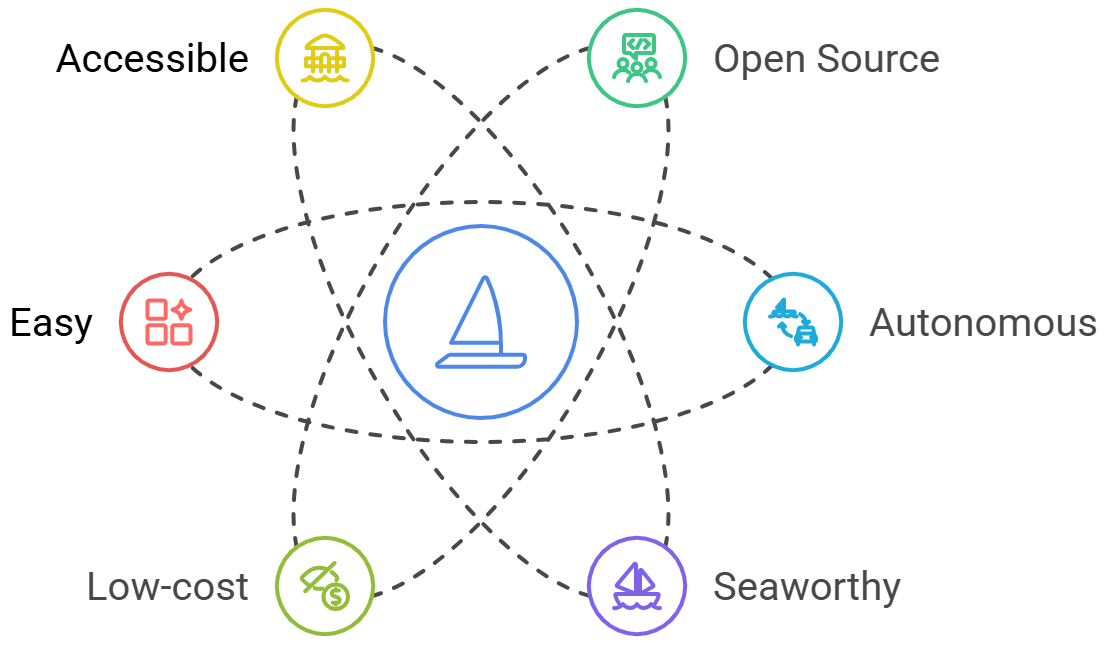}
\caption{Core design principles guiding the development of MARVEL: accessibility, ease of use, low cost, seaworthiness, autonomy, and open-source compatibility.}
\label{fig:system_block_diagram}
\end{figure}

As illustrated in Figure~\ref{fig:system_block_diagram}, the development was guided by six core principles: accessibility, ease of use, low cost, seaworthiness, autonomy, and open-source compatibility. While seaworthiness and autonomy were key technical targets, the platform was also intentionally designed to be accessible, easy to operate, and low-cost to ensure it could serve a wide range of users and research institutions. Accessibility was addressed by using widely available hardware, modular open-source software, and clear documentation, allowing the system to be operated and adapted by students and early-career researchers. Ease of use was achieved by simplifying setup, calibration, and control processes, streamlining field deployment even for users with limited technical backgrounds. The low-cost approach relied on off-the-shelf components, reused lab materials, and open-source tools, resulting in a robust yet budget-conscious solution.

The primary goal was to design, build, and validate a cost-effective, fully autonomous USV capable of operating in diverse marine environments while meeting specific operational requirements. The USV needed to be seaworthy, capable of maintaining stability in wave heights up to \SI{2}{\meter}, while sustaining a steady course at a minimum speed of \SI{2}{\knot}, even under wind conditions reaching \SI{20}{\knot}. Portability was another fundamental consideration, requiring the system to be easily handled by a two-person team for seamless transportation, deployment, maintenance, and retrieval in field experiments. 

Sensor integration was at the core of MARVEL’s design, allowing for the seamless inclusion of multiple navigational and environmental sensors. 

Ensuring high maneuverability was a critical design objective, requiring precise control under dynamic maritime conditions. Differential thrust control, coupled with advanced navigation algorithms, enabled station-keeping capabilities and precise path tracking. Cost optimization was also prioritized throughout the project, leveraging pre-existing hardware components and minimizing financial investment by repurposing available resources. This included the use of commercial off-the-shelf (COTS) sensors, waterproof housings, Raspberry Pi computing units, and Pixhawk controllers, all selected to ensure performance while remaining budget-conscious. To further support cost efficiency, open-source frameworks such as ArduPilot were utilized for navigation, control, and data processing. Beyond cost-effectiveness, ease of use and accessibility were key design considerations. MARVEL was built to be operated, maintained, and adapted by users with varying levels of technical expertise, including students and early-career researchers. Setup, deployment, and control workflows were streamlined to reduce operational complexity, while comprehensive documentation and modular architecture promote accessibility and adaptability across diverse research needs.

In addition to autonomy and cost-efficiency, MARVEL was designed with an emphasis on safety and robustness. The system incorporated fail-safe mechanisms, including emergency recovery procedures, fire hazard mitigation, thermal management, and redundant communication links to ensure reliability in varied marine conditions. The USV’s power system was optimized to support extended mission endurance, with the onboard battery configuration designed to sustain operations for at least two hours. Communication infrastructure included short-range WiFi and RF telemetry, enabling real-time control and data transmission while ensuring continuous mission monitoring.  

The methodology employed for MARVEL’s development followed a structured approach, progressing through conceptual design, structural and mechanical optimization, power and communication system implementation, electronics and sensor integration, autonomous control development, and extensive testing. The initial phase involved defining operational scenarios and environmental constraints while leveraging the available resources within the university’s research facilities. This was followed by mechanical design and fabrication, incorporating hydrodynamic optimizations and weight distribution strategies to enhance stability and efficiency. The electrical and power systems were then developed to ensure stable energy distribution and reliable operation of onboard subsystems. The integration of sensors and computing components was carried out in parallel, ensuring seamless communication between different data acquisition and processing units.

\section{System Design and Architecture} \label{sec:hardware}

This section presents the design and architecture of the MARVEL USV, structured around key subsystems that enable its autonomous capabilities. The design process emphasized modularity, cost-efficiency, and reusability of components. The following subsections outline the structure, power distribution, propulsion mechanisms, onboard sensors, communication architecture, and software systems that constitute the MARVEL platform.

\subsection{Structure}

\begin{figure}[htp]
\centering
\includegraphics[width=0.5\textwidth]{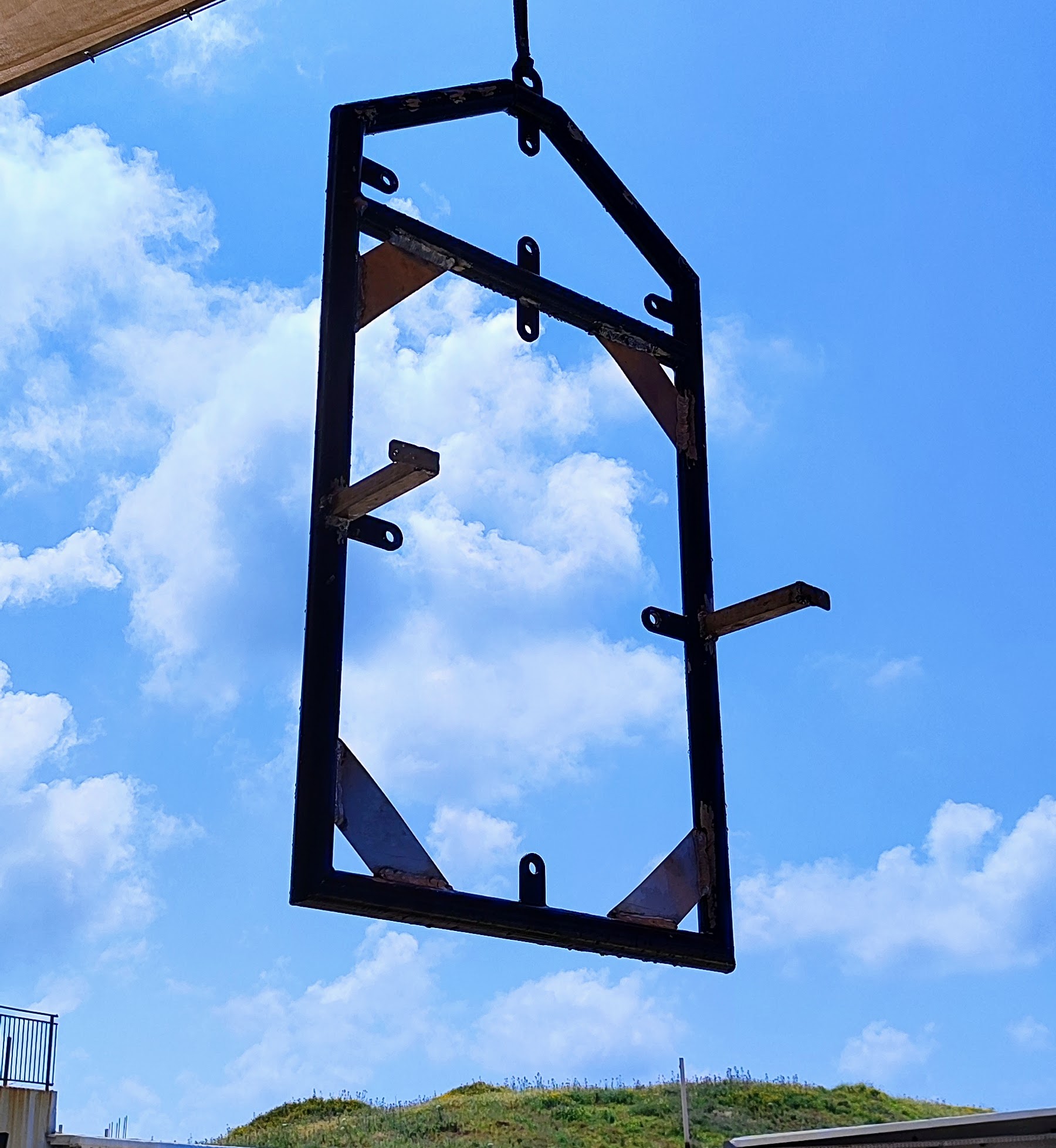}
\caption{Main structural frame of MARVEL following welding and paint application, left to dry before assembly.}
\label{fig:deployment4}
\end{figure}

The design of MARVEL was driven by a combination of operational constraints, cost limitations, and available resources, requiring careful decision-making to balance performance, affordability, and adaptability. As stated in Section II, a primary constraint was ensuring that the USV remained seaworthy while leveraging existing materials to minimize costs. This led to the use of a pre-existing flat aluminum chassis, which was modified rather than designed from scratch. The flat-bottom structure provided a stable platform for sensor integration and was reinforced to improve durability and hydrodynamic performance. Figure \ref{fig:deployment4} shows the main structural frame of MARVEL drying after welding and paint application, captured prior to the final assembly phase. To address the stability challenges of a flat chassis in rough waters, optional side floats were incorporated into the design, allowing for additional buoyancy and lateral stability when required, while keeping the base configuration lightweight for easy transport. Custom 3D-printed brackets and Delrin (POM) components were used to secure onboard electronics and sensors, reducing vibration effects without adding excessive weight. 

The integration of multiple sensors and electronic systems required careful consideration of waterproofing. Rather than sourcing new waterproof enclosures, pre-existing waterproof casings were repurposed and modified to accommodate MARVEL’s unique sensor configuration. These enclosures were reinforced with depth-rated subsea connectors, which provided secure power and data transmission while preventing water ingress. To optimize thermal dissipation, the housings were designed to allow direct contact with water, leveraging passive cooling instead of requiring additional thermal management components. \\
\\

\subsection{Propulsion}

The propulsion system had to support differential thrust control for enhanced maneuverability while being cost-effective and compatible with existing control architectures. Pre-existing BlueRobotics T200 electric marine thrusters were used, as they provided sufficient thrust without requiring major structural modifications. Their mounting positions were carefully adjusted to ensure optimal thrust alignment, compensating for the effects of lateral drift due to the flat hull design. The thrusters were directly integrated with the Pixhawk 6C autopilot, ensuring precise real-time control and adaptability for autonomous operations.

\subsection{Power}

The USV’s power system was designed to balance endurance and energy efficiency while maintaining a low-cost approach. The decision to use a 12V 14Ah lead-acid battery was driven by availability and robustness. Lead-acid batteries, although heavier than lithium-ion alternatives, were already available in the research facility, and reduced overall project costs. The electrical system was structured with separate power circuits for propulsion and onboard electronics to prevent power fluctuations from affecting sensor operation. Additional step-down voltage regulators were used to provide stable 5V supplies for low-power components.

\subsection{Control System}

The choice of Pixhawk 6C as the primary control unit was dictated by its open-source nature, proven reliability in marine applications, off the shelf availability, affordable price and compatibility with ArduPilot. As an industry-standard autopilot, it provided robust navigation, guidance, and control capabilities while allowing for custom software integration. Since computational flexibility was necessary for handling real-time sensor fusion and mission planning, a Raspberry Pi 4 B+ was selected as the mission computer. This decision was influenced by its low cost, high availability, and ease of integration with ArduPilot. While more powerful alternatives existed, the Raspberry Pi provided sufficient processing power for data logging and synchronisation , real-time communication, and auxiliary sensor processing without introducing excessive energy demands.

\subsection{Communication}

\begin{figure}[ht]
\centering
\includegraphics[width=0.45\textwidth]{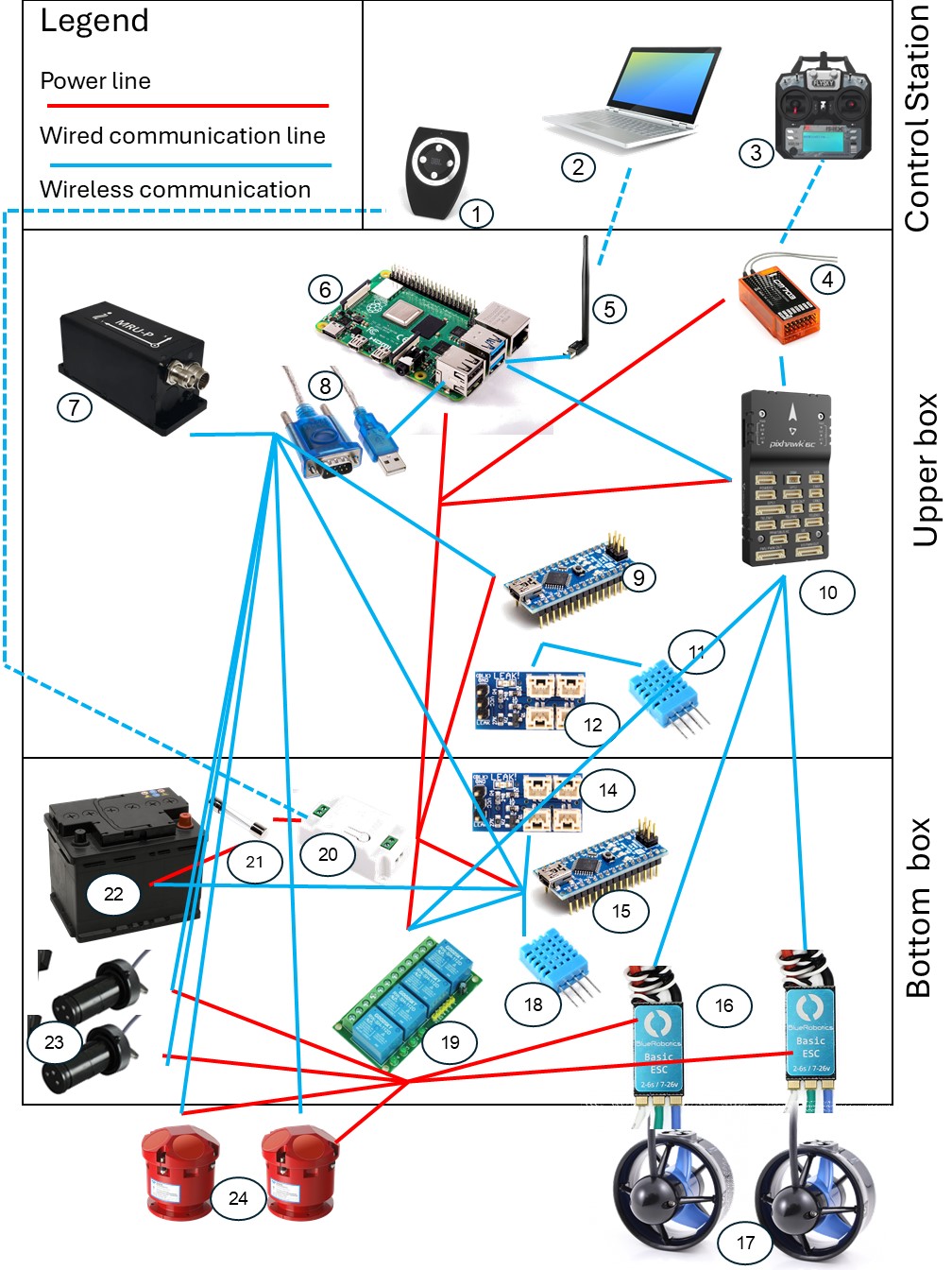}
\caption{System overview diagram showing power, data, and wireless communication lines between major components of the MARVEL USV.}
\label{fig:system_overview2}
\end{figure}

The communication architecture needed to support both real-time telemetry and remote control capabilities, ensuring reliable operation even in GNSS-denied environments. A WiFi-based telemetry link was selected for high-bandwidth communication with the ground station, enabling live mission monitoring and sensor data streaming. Additionally, a low-frequency RF link was integrated for backup telemetry, ensuring basic control retention in case of connectivity loss. The communication system was further supplemented by an AT10 remote control module, allowing for direct manual operation if required. Figure \ref{fig:system_overview2} provides a comprehensive schematic of MARVEL’s system architecture, organized into three distinct segments: the control station, the upper box, and the bottom box. The control station includes an RC switch controller (1), a laptop (2) for mission planning and remote debugging, and a radio controller (3) for manual overrides. The RC transmitter communicates with the onboard receiver (4), while WiFi communication between the laptop (2) and the onboard Raspberry Pi (6) is handled via a wireless dongle (5). All sensor data are routed to the Raspberry Pi (6) through serial-to-USB converters (8), with the MRU-P unit (7) directly connected and independently powered. Each of the two onboard boxes contains an Arduino microcontroller (9 and 15), equipped with environmental (temperature, pressure, humidity) sensors (11 and 18) and leak detectors (12 and 14). A main switch (20) connects the 12V lead-acid battery (22) to a fused split circuit (21), separating power between propulsion and electronics. The Pixhawk flight computer (10) controls the electronic speed controllers (ESCs, 16), which drive the marine thrusters (17), and toggles the power to the DVLs (24) via relays (19). Both DVLs (24) and EM logs (23) communicate via serial with the Pi (6) and draw power from the electronics bus. This setup enables modular, fault-tolerant communication, power distribution, and data logging under real-time mission scenarios.

\subsection{Sensors}

A major design objective was enabling real-time sensor fusion experiments, requiring seamless integration of multiple navigation, velocity, and environmental sensors. The choice of specific sensors was influenced by their availability and the need to support redundant fusion strategies for comparative evaluation. MARVEL was therefore equipped with two Doppler velocity logs (the Waterlinked A50 and the ROWE SeaPILOT), two electromagnetic speed logs (the DX900+ AIRMAR and the Ben Alize EM log), multiple inertial measurement units, and an MRU-P RTK-INS, which served both as a high-accuracy positioning system and a ground-truth reference during testing. 

To support autonomy and real-time data acquisition, the Pixhawk 6C was used as the primary autopilot, handling navigation and low-level control, while a Raspberry Pi 4 B+ acted as the onboard processing unit responsible for sensor communication, logging, and time synchronization. All serial-based sensors were connected through the Raspberry Pi, while the Pixhawk managed critical control-loop sensors, ensuring a clean separation between high-level processing and real-time control. Additional components included a RadioLink AT10II controller for manual override when needed, and a laptop-based ground station used for mission planning and live telemetry monitoring.

\subsection{Software}

\begin{figure}[ht]
\centering
\includegraphics[width=0.45\textwidth]{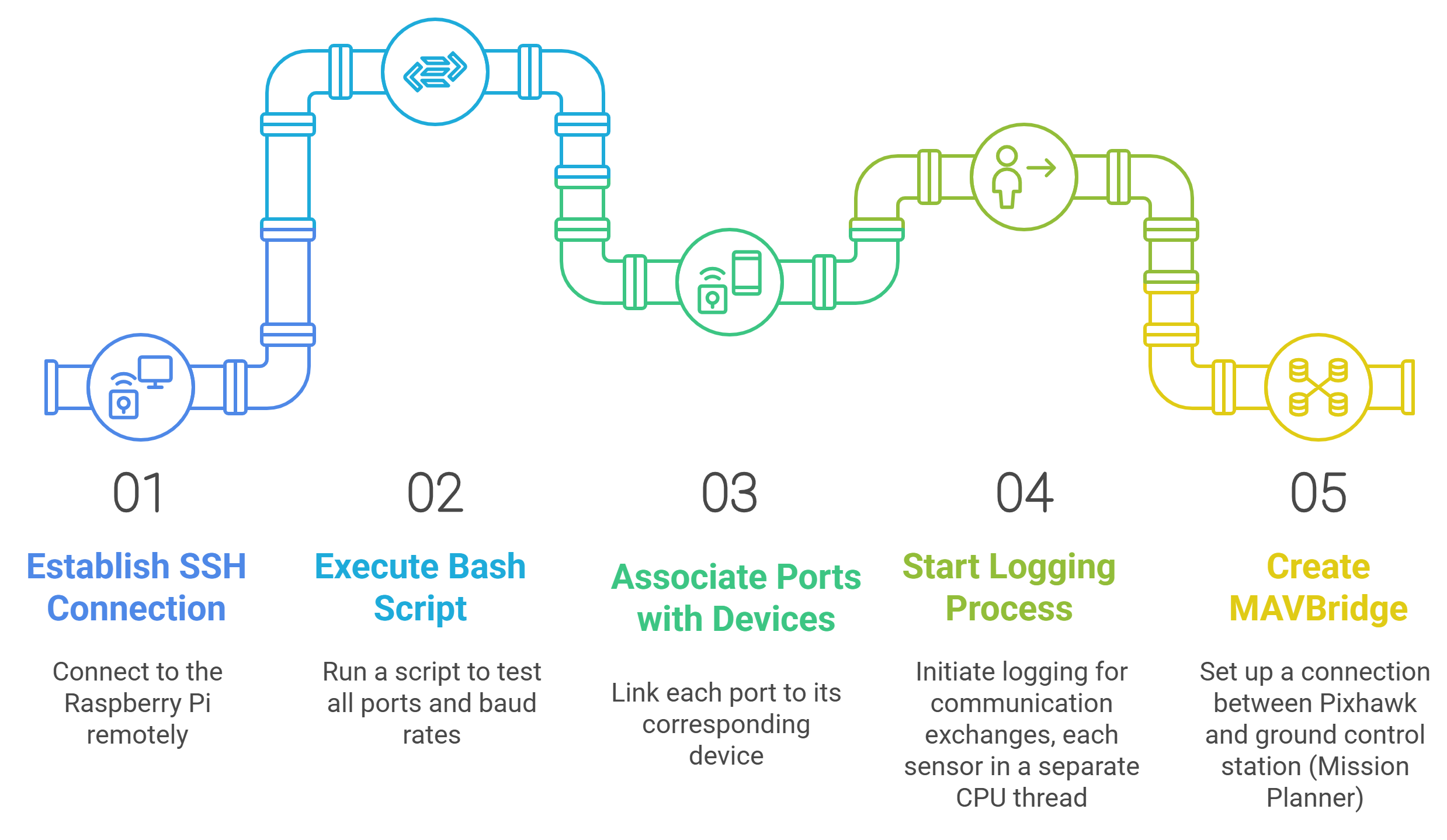}
\caption{Software workflow in the onboard computer.}
\label{fig:software_workflow}
\end{figure}

Figure~\ref{fig:software_workflow} illustrates the software workflow implemented on the onboard computer, enabling systematic port-device association, synchronized multi-sensor logging, and real-time data flow while allowing a MAVLink bridge to the ground control station.

\subsection{Safety}

Given the autonomous nature of MARVEL, robust fail-safe mechanisms were essential to prevent system failures from compromising mission execution. To this end, several safety measures were integrated. Redundant power monitoring was implemented to allow for real-time battery health tracking, preventing unexpected power failures. Emergency motor shutdown protocols were established, allowing for manual and automated system overrides in case of hardware malfunction. A visual signaling system was included to improve operational safety during nighttime and low-visibility conditions. A radar reflector was incorporated to enhance MARVEL’s detectability by nearby vessels, reducing collision risk during autonomous deployments.

\subsection{Summary}

The MARVEL USV represents a highly adaptable, low-cost research platform that supports real-world sensor fusion validation. By leveraging existing resources while incorporating scalable modularity, the platform remains flexible for future hardware upgrades and software enhancements. The internal layout and available mounting points leave room for integrating additional navigation, environmental, or perception sensors, making MARVEL an evolving testbed for ongoing and future research in autonomous marine systems. The next phase of development will focus on extending operational endurance, and enhancing autonomous capabilities for dynamic mission execution.

\section{Assembly and  Experiments} \label{sec:assembly_testing}

The following section presents the multi-stage assembly and validation process for the MARVEL USV, from mechanical construction through to real-world deployment. Testing was conducted in stages, allowing for iterative improvements based on empirical feedback.

\subsection{Mechanical and Electrical Assembly}

The construction of MARVEL followed a rigorous, multi-stage process to ensure that every structural, electronic, and software component functioned as intended. Given the scope of work, progress required careful time management, balancing hands-on assembly, debugging, and testing with existing professional commitments. Each stage, from mechanical fabrication to real-world trials, was carried out with a methodical approach to maximize efficiency while accommodating the constraints of a limited working schedule.

\begin{figure}[hbp]
\centering
\includegraphics[width=0.5\textwidth]{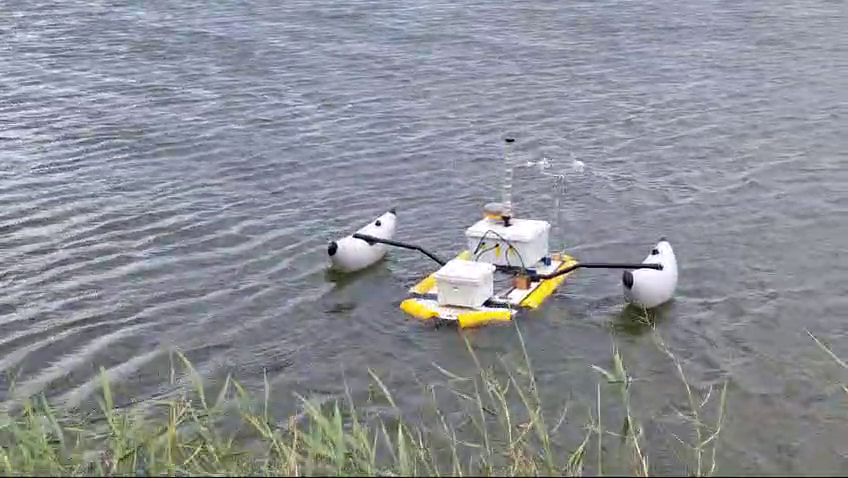}
\caption{Field deployment of MARVEL under strong wind conditions.}
\label{fig:deployment2}
\end{figure}

The assembly began with structural modifications to adapt the pre-existing aluminum chassis, reinforcing it to accommodate the planned payload and operational loads. Extensive stick welding was required to mount brackets, reinforcements, and supports for propulsion, sensor housings, and power units. The waterproof electronic enclosures were fitted with custom-designed outer structures made of Delrin (POM), providing a stable and vibration-resistant foundation for securing batteries, controllers, and processing units. Fixation points for each sensor were fabricated and adjusted to ensure reliable placement and minimal obstruction to data acquisition. Additional work was dedicated to integrating safety features, such as a visual signaling lantern system and a radar reflector, to enhance operational visibility. Figure~\ref{fig:deployment2} shows an early deployment of MARVEL under strong wind conditions with side floats attached, highlighting its seaworthiness during adverse weather. Once structural modifications were completed, protective coatings were applied to reduce corrosion risks, and final adjustments ensured that all mechanical elements were properly aligned before moving to the next phase.

\begin{figure}[H]
\centering
\includegraphics[width=0.45\textwidth]{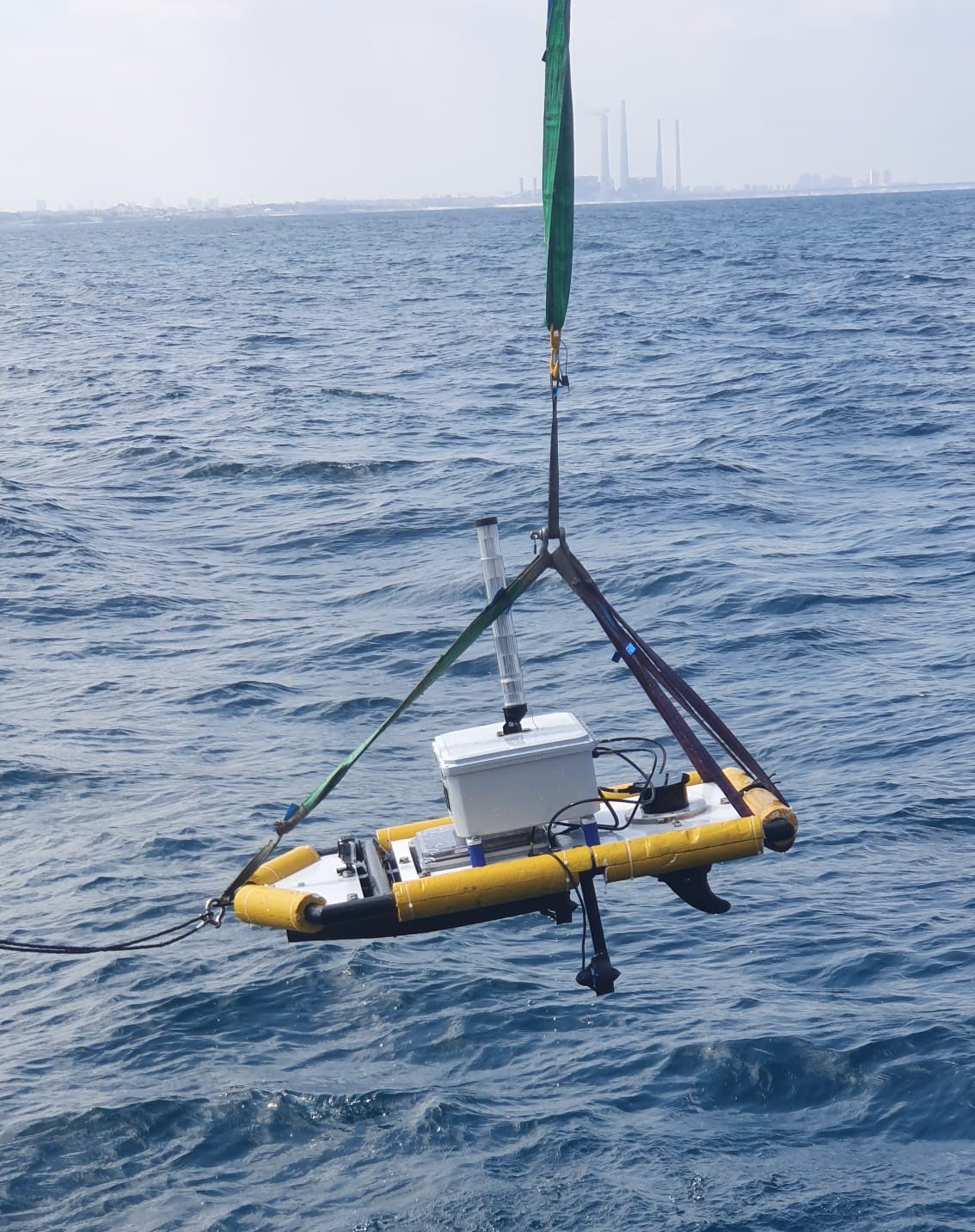}
\caption{MARVEL with sensor payload and electronics compartments in the first sea trials.}
\label{fig:system_overview}
\end{figure}

The electrical integration process involved extensive wiring, requiring numerous cable welds, insulation applications, and routing adjustments to accommodate the USV’s complex sensor suite and propulsion system. To ensure system reliability, independent power circuits were implemented for propulsion, sensor operation, and communication, minimizing electromagnetic interference. Redundant fuses and power monitoring sensors were installed to safeguard against potential failures. With wiring in place, all electronic components, including the Pixhawk 6C autopilot, Raspberry Pi 4 B+ mission computer, additional microcontrollers, and sensor interfaces, were carefully mounted within the waterproof enclosures. Given the intricate nature of this phase, ensuring neat and efficient cable management was essential to maintain accessibility for future maintenance. Figure \ref{fig:system_overview} shows MARVEL during its initial sea trials, including visible sensor payloads and electronic compartments.

\subsection{Software Integration and Debugging}

Once the physical assembly was completed, the next stage focused on software development, configuring the Pixhawk, Raspberry Pi, and microcontrollers to enable autonomous navigation, sensor fusion, real-time data logging, and remote communication. Developing and debugging the necessary scripts required significant effort, with careful testing of sensor synchronization and log integrity. The communication system was set up with both WiFi telemetry and an RF backup link, ensuring continuous data transmission to the ground station. Particular attention was given to logging efficiency, refining data formats, and ensuring all sensor readings were time-stamped correctly for post-mission analysis.

\subsection{Testing and Field Validation}

Once hardware and software integration was complete, system debugging and dry testing were performed to ensure reliability before full deployment. Mechanical and electrical checks validated structural integrity and power stability, while software testing focused on verifying telemetry, logging, and navigation commands under simulated conditions. The complete system was then evaluated in land-based scenarios, emphasizing sensor synchronization, data consistency, and calibration accuracy across all subsystems.

Once dry tests confirmed the system’s functionality, waterproofing validation was conducted. Each electronic housing was submerged and subjected to pressure tests to verify its resistance to water ingress. Internal monitoring of humidity, temperature, and pressure within the enclosures provided additional confirmation of their reliability over prolonged exposure. Any detected vulnerabilities were addressed before proceeding to in-water trials. 

With structural, electrical, and software systems fully operational, MARVEL underwent its first water trials in a controlled pool environment. These tests focused on maneuverability, propulsion efficiency, and stability, allowing for early-stage adjustments. Differential thrust responses were tuned, and further refinements were made to the calibration of velocity measurement sensors, particularly the DVL and EM logs. Observations from these tests led to minor modifications, including adjustments to motor mounting angles to optimize thrust alignment.

Following successful pool trials, MARVEL was deployed in calm coastal waters to validate navigation, stability, and endurance. A sequence of autonomous missions was conducted to assess its ability to maintain predefined waypoints, resist environmental disturbances, and sustain consistent velocity estimates. Initial trials in calm conditions confirmed the correct functioning of navigation algorithms, heading stability, and sensor fusion. 

\begin{figure}[hb]
\centering
\includegraphics[width=0.5\textwidth]{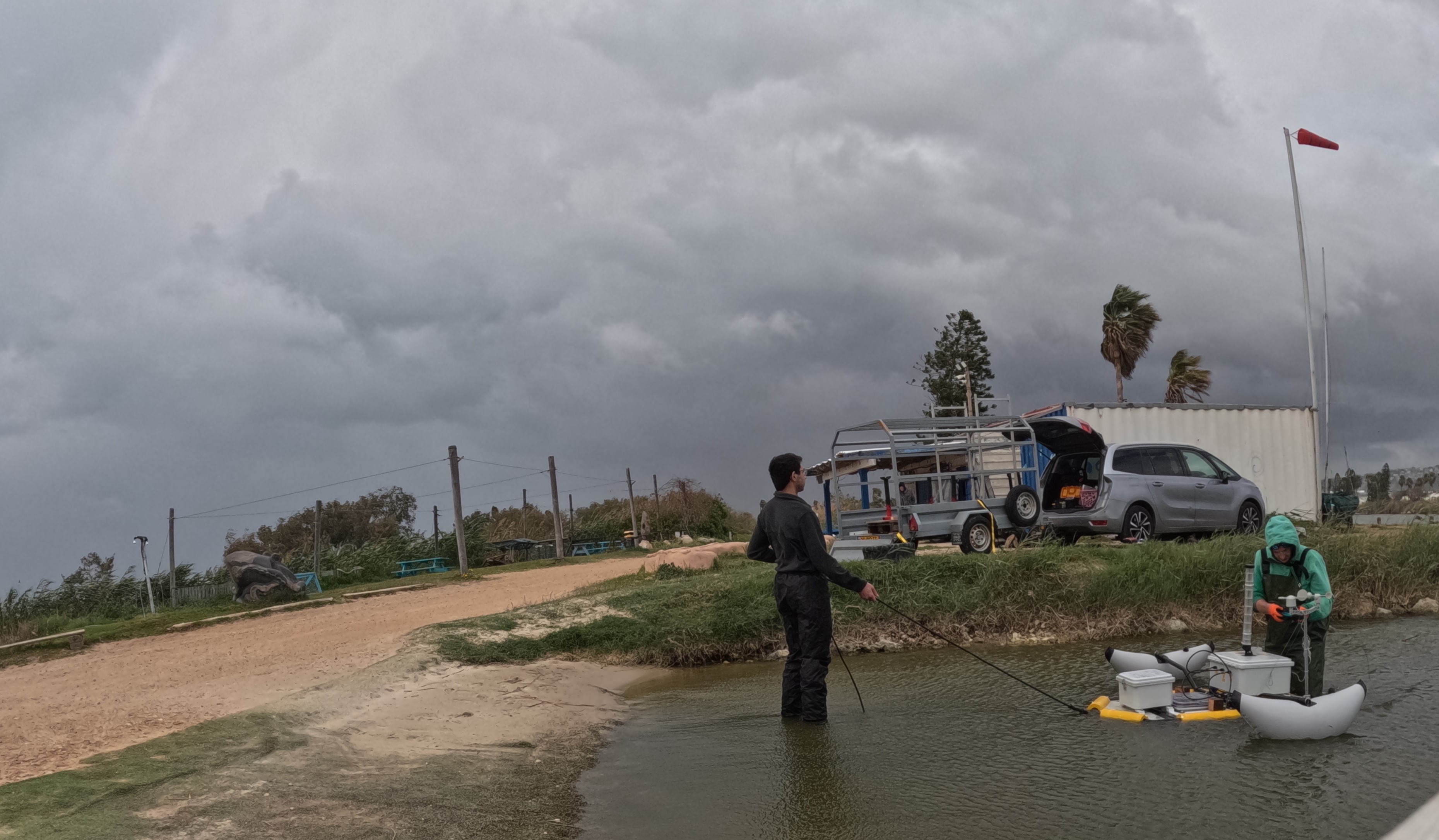}
\caption{Field deployment of MARVEL under strong wind conditions.}
\label{fig:deployment3}
\end{figure}

After confirming basic performance in controlled conditions, MARVEL underwent testing in more demanding real-world scenarios, including GNSS-denied environments. The response of the control system in higher sea states was also tested, assessing MARVEL’s behavior under increased wave and current loads. Optional side floats were attached and removed during different trials to measure their impact on stability in varying conditions. Figure \ref{fig:deployment3} shows another field deployment of MARVEL in strong wind conditions, highlighting operational testing in dynamic maritime environments.

Each round of testing led to iterative refinements in both hardware and software. Adjustments were made to motor control parameters, and data filtering strategies to improve overall accuracy. Power consumption analysis suggested optimizations in propulsion efficiency, contributing to extended mission endurance. These findings helped guide further modifications, ensuring that MARVEL could operate with greater reliability in prolonged deployments.

Beyond system validation, it was essential to document operational procedures for future use. A detailed guide was prepared covering mission planning, pre-deployment checks, calibration workflows, real-time monitoring, and post-mission data analysis. This documentation aimed to ensure consistent and repeatable deployments while providing a reference for troubleshooting and future system expansions.

The completion of MARVEL’s development through this extensive process demonstrated the viability of a cost-effective, modular USV capable of real-world autonomous navigation and sensor fusion research. Managing the assembly, testing, and iterative improvements alongside other professional responsibilities required careful coordination and persistence, reinforcing the practical challenges of building a fully functional research platform within time human resource and budget constraints.

\section{Conclusion} \label{sec:conclusion}
This paper presented the design, construction, and validation of MARVEL—an autonomous unmanned surface vehicle developed as a modular research platform for real-world testing of sensor fusion and navigation algorithms, particularly in GNSS-denied environments. The platform was designed under strict constraints of low cost, portability, and seaworthiness, while adhering to core principles including autonomy, open-source compatibility, accessibility, and ease of use.

Through the integration of redundant velocity and inertial sensors—such as DVLs, EM logs, IMUs, and a GNSS/RTK system—MARVEL enables synchronized high-frequency data acquisition and robust autonomous navigation. The system architecture supports modular expansion and has demonstrated strong performance during field deployments across varying sea conditions.

MARVEL represents a novel contribution to the field by providing an affordable, accessible, and fully autonomous USV for validating advanced navigation techniques in real-world conditions. Unlike typical solutions that rely on simulation or single-sensor setups, MARVEL supports multi-sensor fusion experiments with direct and accurate ground truth.

Future work will focus on extending MARVEL’s capabilities to include onboard deep learning for adaptive navigation, real-time decision-making, and intelligent mission planning. The platform will also be used to collect large-scale maritime datasets, supporting the development and validation of AI-driven marine autonomy frameworks.

\section*{Acknowledgments}
The authors are grateful to Itai Savin, for his continued support in several sea experiments and MARVEL's mechanisms.

\bibliographystyle{IEEEtran}
\bibliography{export}

\end{document}